\documentclass[letterpaper]{article} 
\usepackage{aaai2027}  
\usepackage[hyphens]{url}  
\usepackage{graphicx} 
\usepackage{natbib}  
\usepackage{caption} 
\usepackage{algorithm}
\usepackage{algorithmic}

\usepackage{newfloat}
\usepackage{listings}
\DeclareCaptionStyle{ruled}{labelfont=normalfont,labelsep=colon,strut=off} 
\floatstyle{ruled}
\newfloat{listing}{tb}{lst}{}
\floatname{listing}{Listing}

\usepackage{booktabs}

\usepackage{amsmath}
\usepackage{amssymb}

\usepackage{pifont}
\usepackage{makecell}

\title{GaussianSeed: Hierarchical Gaussian Seeding for High-Resolution \\ 3D Occupancy Prediction}
\author{
    Xinzhuo Li\equalcontrib,
    Xianghui Pan\equalcontrib,
    Jiayuan Du,
    Wei Wei,
    Liuyi Wang,
    Chengju Liu\corresponding,
    Qijun Chen
}
\affiliations{

    Tongji University, Shanghai, China \\
    \{2531772, 2111119, 2011239, 2510856, wly, liuchengju, qjchen\}@tongji.edu.cn
}

\begin{document}

\maketitle

\begin{abstract}
Vision-centric 3D occupancy prediction provides dense scene representations essential for autonomous driving and robotic navigation, yet existing methods struggle to scale to high voxel resolutions due to prohibitive computational costs. To address this, we introduce GaussianSeed, a progressive multi-scale Gaussian occupancy prediction framework that organizes primitives into a coarse-to-fine hierarchy. Benefiting from this hierarchical design, GaussianSeed effectively circumvents the memory bottlenecks inherent in dense representations, successfully scaling to a $0.1\text{m}$ spatial resolution while maintaining real-time inference capabilities. To comprehensively evaluate high-resolution geometric perception, we further construct TJScenes, a panoramic six-camera occupancy dataset with highly detailed $0.1\text{m}$ annotations. Extensive experiments on Occ3D-nuScenes and TJScenes demonstrate that GaussianSeed delivers the lowest latency among all evaluated methods while maintaining highly competitive accuracy, advancing the efficiency-quality frontier of high-resolution 3D occupancy prediction.
\end{abstract}


\section{Introduction}

\begin{figure}[t]
    \centering
    \includegraphics[width=\columnwidth]{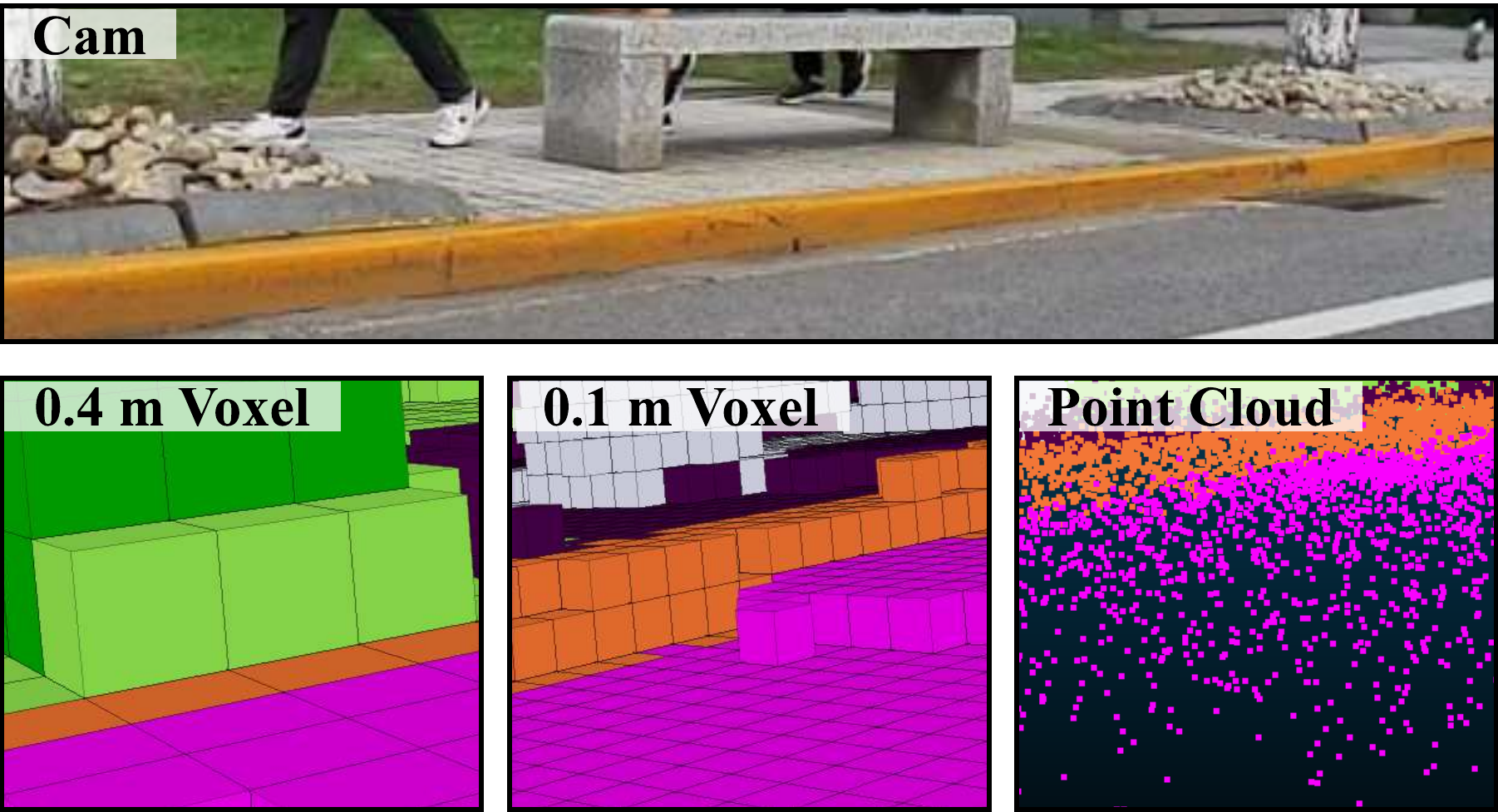} 
    \caption{\textbf{The same scene represented in different resolutions.} Voxels of $0.4\text{m}$ fall short to describe the curbs clearly, while a fine-grained resolution of $0.1\text{m}$ can accurately preserve such subtle topological boundaries.}
    \label{fig:teaser_04m01m}
\end{figure}

Recent years have witnessed significant advances in vision-centric 3D occupancy prediction~\cite{zhang2024visionbased3doccupancyprediction}. However, existing methods predominantly rely on dense 3D feature representations, which introduce unnecessary computational overhead and struggle to scale to higher resolutions required for detailed scene understanding. Although several pioneering works have turned to sparse representations for better scalability, they still fall short when scaling to the fine-grained resolutions required for intricate scene understanding. As illustrated in Fig.~\ref{fig:teaser_04m01m}, promoting the grid grain from a coarse scale to a high-fidelity $0.1\text{m}$ resolution allows the system to clearly discern subtle road topologies like curbs, which is crucial for downstream vehicle planning. Generally, existing sparse designs either employ hierarchical voxel pruning or formulate the task as sparse point set prediction. The former relies heavily on early-stage coarse predictions, making it prone to irreversible information loss where fine-grained, small objects are easily pruned as background by mistake. The latter bypasses space grids via point coordinate regression, yet the number of required points escalates cubically with higher spatial resolutions, leading to prohibitive computational overhead in query matching and attention mechanisms. Consequently, existing sparse designs still fail to achieve a satisfactory trade-off between high resolution and high efficiency.

\begin{figure*}[ht]
    \centering
    \includegraphics[width=\textwidth]{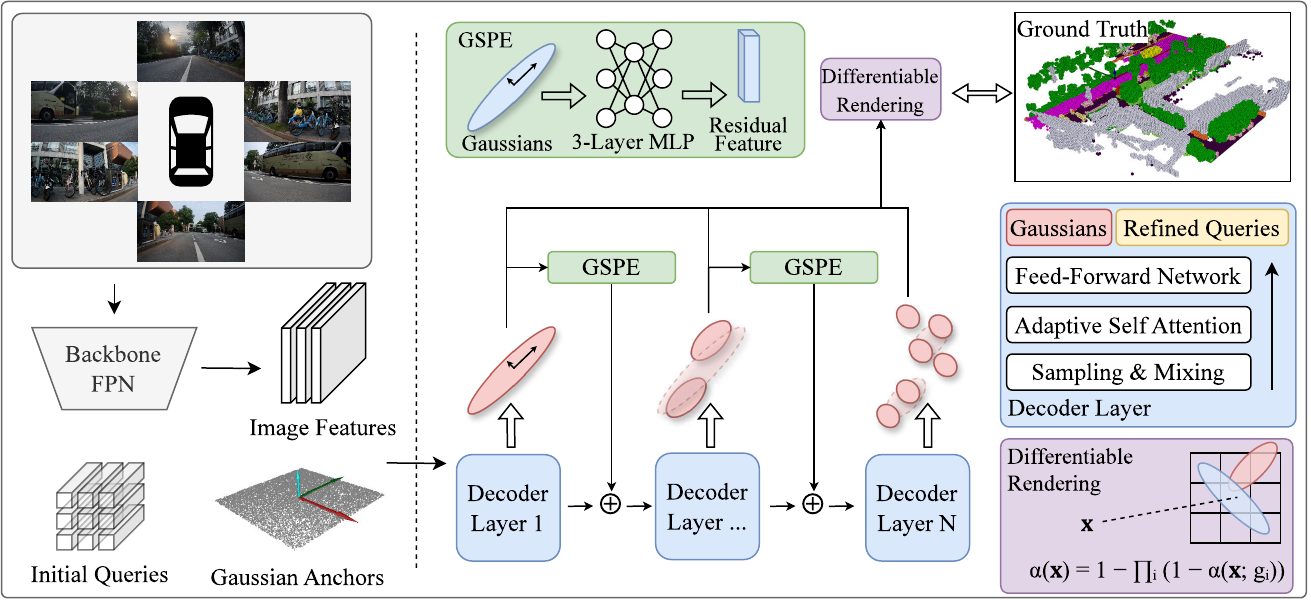} 
    \caption{\textbf{The overall architecture of the GaussianSeed framework.} At each decoder layer, queries are progressively refined leveraging multi-scale image features. These refined queries are subsequently decoded into 3D Gaussian primitives and rendered into the voxel space, where the dense occupancy ground truth is applied for direct supervision.}
    \label{fig:framework}
\end{figure*}

A further challenge is the limited diversity of publicly available occupancy benchmarks. The widely adopted Occ3D-nuScenes~\cite{tian2023occ3d} dataset provides annotations at a relatively coarse resolution of $0.4\text{m}$, masking the scalability issues that emerge at finer granularity. The Waymo Open Dataset~\cite{sun2020waymo}, while offering more detailed geometry, is restricted to five cameras with no rear-facing sensor, precluding full 360-degree scene understanding and limiting its utility for holistic occupancy modeling. There remains a clear need for benchmarks that stress-test both the representational fidelity and the rendering efficiency of occupancy predictors at high voxel resolutions.

To bridge these gaps, we introduce GaussianSeed, a progressive multi-scale Gaussian occupancy framework that adaptively scales to high resolutions without prohibitive costs, and construct TJScenes, a occupancy benchmark with fine-grained $0.1\text{m}$ annotations. GaussianSeed integrates three synergistic components: (i) Regression-Based Gaussian Initialization (RBGI), which anchors Gaussian centers via lightweight coordinate regression without depth priors; (ii) Hierarchical Gaussian Seed Devolution (HGSD), which propagates primitives from coarse to fine across decoder layers; and (iii) Gaussian Seed Parameter Encoder (GSPE), which injects decoded properties into subsequent layers for stable cross-layer refinement. To summarize, our primary contributions are three-fold:
\begin{itemize}
    \item We introduce a multi-scale Gaussian occupancy framework that organizes primitives into a coarse-to-fine hierarchy, elegantly breaking the cubic complexity bottleneck of high-resolution scene understanding.
    \item We propose RBGI, HGSD, and GSPE, three synergistic techniques that together form a camera-only, end-to-end pipeline achieving efficient high-resolution occupancy prediction.
    \item We construct TJScenes, a panoramic 6-camera occupancy dataset annotated at $0.1\text{m}$ resolution. Designed for mobile robotic navigation, TJScenes incorporates diverse off-road and sidewalk scenarios, complementing existing road-centric benchmarks and serving as an ideal testbed for high-fidelity geometric perception.
\end{itemize}

\section{Related Work}
\paragraph{Vision-centric 3D Occupancy Prediction.} 
3D occupancy prediction discretizes a scene into semantic voxel grids, providing a dense holistic representation essential for safe motion planning. Vision-centric approaches dominate this field, typically lifting 2D image features into 3D space via depth-guided projection~\cite{chen2025alocc,monoscene2022,pan2024generocc} or cross-attention~\cite{huang2023tpvformer}, followed by dense 3D convolutions or Transformers for volumetric reasoning. Recent works~\cite{li2023fbocc,gsdocc2025he,protoocc2025kim} adopt BEV features to reduce voxel-level overhead, yet remain constrained by dense paradigms that overlook the inherent spatial sparsity of real-world scenes, leading to unfavorable efficiency-quality trade-offs. 

\paragraph{Sparse Representations for 3D Scene Understanding.}
Sparse representations have recently demonstrated remarkable success across various 3D perception tasks~\cite{du2026sparseworld_tc}. To exploit spatial sparsity, SparseOcc~\cite{liu2023sparseocc} employs hierarchical voxel pruning, but suffers from irreversible early-stage false negatives. OPUS~\cite{wang2024opus} reformulates occupancy as point set prediction, yet is prone to structural hole artifacts and constrained by fixed point budgets. Inspired by 3D Gaussian Splatting~\cite{kerbl3Dgaussians}, GaussianFormer~\cite{huang2024gsformer1} pioneers Gaussian-to-voxel feature rendering for sparse occupancy prediction. GaussianFormer-2~\cite{huang2024gsformer2} further adopts a probabilistic model to reduce primitive redundancy, but requires a separate LSS-based~\cite{philion2020LSS} initialization stage that hinders end-to-end optimization. Moreover, both GaussianFormer variants adhere to a rigid one-to-one query-to-Gaussian assignment, limiting cross-Gaussian interaction to local sparse convolutions and preventing global information propagation.

\paragraph{3D Occupancy Prediction Benchmarks.}
The advance of 3D occupancy prediction relies critically on large-scale, high-quality annotated benchmarks. Occ3D-nuScenes and Occ3D-Waymo~\cite{tian2023occ3d} provide multi-view coverage under diverse weather conditions, establishing standard frameworks for outdoor occupancy evaluation. SSCBench~\cite{li2024sscbench} adapts semantic scene completion to autonomous driving, enabling unified cross-dataset comparisons. To address label sparsity from LiDAR frame superimposition, OpenOccupancy~\cite{wang2023openoccupancy} introduces an interactive Augmenting and Purifying pipeline with extensive human labeling, achieving denser annotations at $0.2\text{m}$ resolution. Despite these efforts, existing benchmarks remain limited in two key aspects: they operate at relatively coarse voxel resolutions ($0.4\text{m}$ or $0.2\text{m}$), failing to capture thin structures and fine boundaries, and they predominantly focus on structured roadway scenarios, overlooking off-road settings essential for micro-mobility and robotic navigation.

\section{Method}

\paragraph{Overview.} As illustrated in Fig.~\ref{fig:framework}, GaussianSeed is a coarse-to-fine architecture that predicts a set of semantic 3D Gaussian primitives $\mathcal{G} = \{g_i\}_{i=1}^M$ from multi-view images. Each Gaussian $g_i$ is parameterized by its mean $\mathbf{m}_i \in \mathbb{R}^3$, scale $\mathbf{s}_i \in \mathbb{R}^3$, rotation quaternion $\mathbf{q}_i \in \mathbb{R}^4$, opacity $a_i \in [0, 1]$, and a semantic logit vector $\mathbf{c}_i \in \mathbb{R}^C$, and these primitives are rendered into dense occupancy grids $\mathbf{O} \in \mathbb{R}^{X \times Y \times Z \times C}$ via a differentiable volumetric pipeline.

Given multi-view images, a 2D backbone with FPN~\cite{lin2017fpn} extracts multi-scale features, which are fed into a cascaded decoder with several refinement layers. Within each layer, 3D queries sample reference points projected onto multi-view image features and aggregate them via Adaptive Mixing~\cite{liu2023sparsebev}. The resulting query features progressively regress Gaussian parameters conditioned on predictions inherited from the preceding layer. To ensure robust convergence across granularities, the Gaussians from each decoder layer are independently rendered and supervised against ground-truth labels.

The Gaussian-to-voxel transformation uses localized feature accumulation, inspired by the probabilistic superposition formulation in GaussianFormer-2~\cite{huang2024gsformer2}. For a voxel coordinate $\mathbf{x} \in \mathbb{R}^3$, a single Gaussian $g_i$ contributes occupancy probability:
\begin{equation}
\label{eq:geo_exp}
\alpha(\mathbf{x}; g_i) = \exp \left( -\frac{1}{2}(\mathbf{x} - \mathbf{m}_i)^T \Sigma_i^{-1} (\mathbf{x} - \mathbf{m}_i) \right),
\end{equation}
where $\mathbf{m}_i$ and $\Sigma_i$ denote the mean and 3D covariance of $g_i$. Following 3D Gaussian Splatting, $\Sigma_i$ is composed of a rotation matrix $\mathbf{R}_i$ (converted from $\mathbf{q}_i$) and a diagonal scale matrix $\mathbf{S}_i = \operatorname{diag}(\mathbf{s}_i)$ via $\Sigma_i = \mathbf{R}_i \mathbf{S}_i^2 \mathbf{R}_i^T$. The holistic occupancy at $\mathbf{x}$ is accumulated via the probability multiplication theorem:
\begin{equation}
\alpha(\mathbf{x}) = 1 - \prod_{i=1}^{P} \left( 1 - \alpha(\mathbf{x}; g_i) \right),
\end{equation}
with $P$ being the number of primitives intersecting voxel $\mathbf{x}$. The semantic logit at $\mathbf{x}$ is a weighted mixture:
\begin{equation}
e(\mathbf{x}; \mathcal{G}) = \frac{\sum_{i=1}^{P} \mathcal{N}(\mathbf{x}; \mathbf{m}_i, \Sigma_i) a_i \tilde{\mathbf{c}}_i}{\sum_{j=1}^{P} \mathcal{N}(\mathbf{x}; \mathbf{m}_j, \Sigma_j) a_j},
\end{equation}
where $\tilde{\mathbf{c}}_i$ is the normalized semantic vector of $g_i$, $a_i$ its opacity, and $\mathcal{N}$ the 3D Gaussian density. The final occupancy prediction combines geometry and semantics, reserving $1-\alpha(\mathbf{x})$ for empty space:
\begin{equation}
\hat{o}(\mathbf{x}; \mathcal{G}) = \left[ 1 - \alpha(\mathbf{x}); \; \alpha(\mathbf{x}) \cdot e(\mathbf{x}; \mathcal{G}) \right].
\end{equation}

\paragraph{Regression-Based Gaussian Initialization (RBGI).}

\begin{figure}[ht]
    \centering
    \includegraphics[width=0.95\columnwidth]{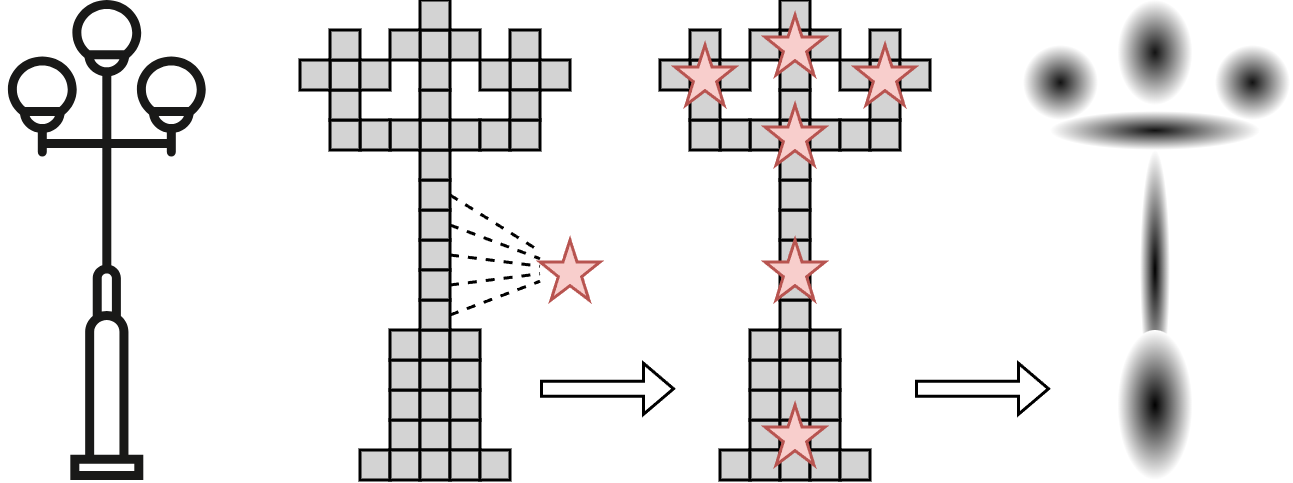} 
    \caption{\textbf{Mechanism of the regression-based Gaussian initialization via Chamfer Distance regularization.} By explicitly initializing the means of the Gaussian primitives to the center points of voxel clusters, this mechanism establishes a stable geometric scaffold for the subsequent optimization of Gaussian parameters.}
    \label{fig:qbased_cdloss}
\end{figure}

As widely acknowledged in multi-view 3D scene understanding, accurate spatial geometry layout serves as a crucial foundation for robust occupancy estimation. For Gaussian-based representations, the spatial influence of each primitive decays exponentially with distance from its center. Consequently, when Gaussians are initialized far from valid geometry, their gradients vanish rapidly, making it intractable to supervise their attributes. Existing methods mitigate this by introducing depth priors via LSS-based lifting~\cite{huang2024gsformer2} or auxiliary LiDAR~\cite{doruk2026gaussianocc3d,zhao2026gaussianformer3d}. To maintain a camera-only, end-to-end pipeline, we propose a lightweight regression paradigm, illustrated in Fig.~\ref{fig:qbased_cdloss}. By decoupling center estimation from downstream attribute prediction, the network first anchors Gaussians into geometrically valid regions, establishing a stable scaffold before synthesizing fine-grained properties.

Formally, within the $l$-th decoder layer, a set of learnable 3D queries $\mathcal{Q}^l = \{\mathbf{q}_i^l\}_{i=1}^Q$ interacts with the multi-view image features via cross-attention to directly regress the coordinate offsets. For each Gaussian belonging to each query, the predicted center $\hat{\mathbf{m}}_{i,(r)}^l \in \mathbb{R}^3$ is formulated as:
\begin{equation}
\label{eq:residual_means_update}
\hat{\mathbf{m}}_{i,(r)}^l = \sigma(W_{pos} \cdot \mathbf{q}_i^l) + \mathbf{a}_i^l,
\end{equation}
where $W_{pos}$ denotes a linear projection matrix, $\sigma(\cdot)$ is the sigmoid activation function, and $\mathbf{a}_i^l$ represents the structural anchor corresponding to the average position of the Gaussians inherited by query $i$ from the preceding layer.

To regularize these positions without explicit depth lifting, we avoid scale-dependent optimal transport (e.g., Sinkhorn~\cite{cuturi2013sinkhorn}) due to the mutual dependency between scale estimation and position optimization, and instead adopt the bidirectional Chamfer Distance ($\mathcal{L}_{cd}^l$) to supervise positional regression at each decoder layer:

\begin{equation}
\label{eq:cdloss}
\mathcal{L}_{cd}^l = \frac{1}{G^l} \sum_{i=1}^{G^l} \min_{j} \|\hat{\mathbf{m}}_i^l - \mathbf{p}_j\|_2^2 + \frac{1}{K} \sum_{j=1}^K \min_{i} \|\hat{\mathbf{m}}_i^l - \mathbf{p}_j\|_2^2,
\end{equation}
where $\mathbf{p}_j$ denotes the $j$-th ground-truth voxel center point, $K$ the total number of such points, and $G^l$ the total number of Gaussians at layer $l$. Since $K \gg G^l$, the second term forces Gaussian centers toward the centroids of local point clusters, consistent with the exponential spatial decay in Eq.~(\ref{eq:geo_exp}). The total CD loss is summed over all decoder layers: $\mathcal{L}_{cd} = \sum_l \mathcal{L}_{cd}^l$.

\paragraph{Hierarchical Gaussian Seed Devolution.}

While the decoupled regression paradigm successfully stabilizes the positional regression within a single decoder layer, scaling this formulation to dense scene representations introduces significant optimization challenges. A straightforward approach would be to predict all the Gaussians in a single step. However, this presents a severely ill-conditioned optimization problem: thousands of Gaussians must simultaneously resolve global scene layout and local geometric details from a cold start, leading to slow convergence and suboptimal local minima. The vast search space and the intricate optimization landscape make it intractable for the architectures to coordinate coarse structural topology and fine-grained geometry concurrently.

\begin{figure}[ht]
    \centering
    \includegraphics[width=0.85\columnwidth]{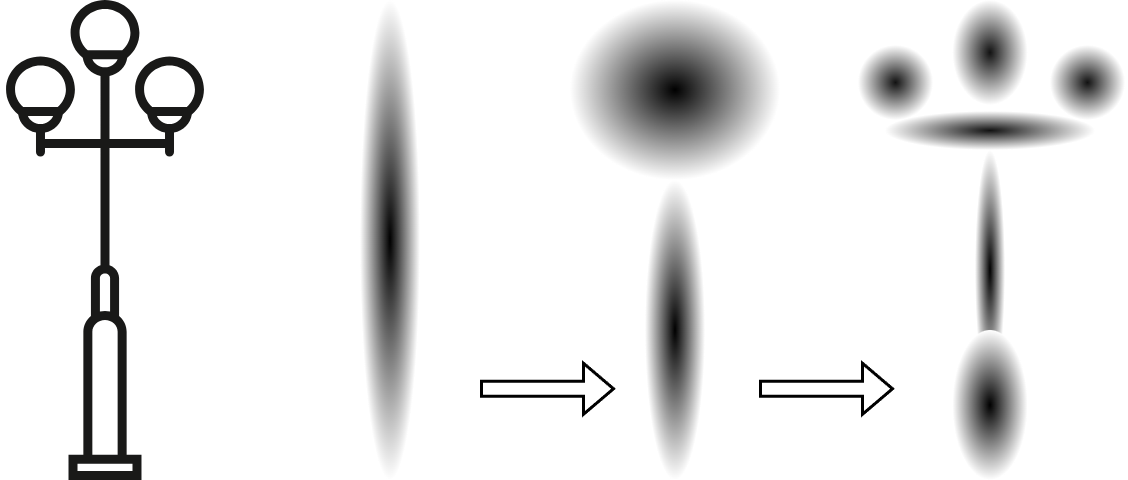}
    \caption{\textbf{Progressive refinement in HGSD.} Early decoder layers anchor large-scale Gaussians onto reliable spatial positions (middle), while later layers utilize finer Gaussians to capture intricate geometric details. }
    \label{fig:hgsd}
\end{figure}

To alleviate this structural bottleneck, we present a Hierarchical Gaussian Seed Devolution (HGSD) strategy, as illustrated in Fig.~\ref{fig:hgsd}. Instead of expanding the query capacity inside a single decoder layer, our core philosophy is to model the Gaussian generation as a progressive, coarse-to-fine propagation process across sequential decoder layers. Concretely, a query $\mathcal{Q}_i$ starts with a single Gaussian ($R_0=1$) that captures the dominant structure within its spatial neighbourhood. As decoding proceeds, this Gaussian devolves into progressively more fine-grained offspring, each specializing in a sub-region via a learned offset $\Delta \mathbf{m}$ relative to the parent anchors ${\mathbf{a}}^{l-1}$:

\begin{equation}
    \mathbf{m}^l = {\mathbf{a}}^{l-1} + \Delta\mathbf{m}^l, \quad {\mathbf{a}}^{l-1} = \frac{1}{R^{l-1}}\sum_{k=1}^{R^{l-1}} \mathbf{m}_{(k)}^{l-1},
\end{equation}
where $\Delta\mathbf{m}^l$ corresponds to the predicted coordinate offset, formulated in Eq.~\ref{eq:residual_means_update}.

In addition to the explicit inheritance of spatial coordinates, we introduce a lightweight Gaussian Parameter Encoder (GSPE). Let $\mathcal{G}^l$ and $\mathcal{C}^l$ denote the predicted geometric attributes and semantic logits from the $l$-th decoder layer. GSPE encodes these into a feature vector via a 3-layer MLP and injects it into the subsequent layer's query embeddings via a residual connection:
\begin{equation}
f_{gs}^l = \text{MLP}\big( [\mathcal{G}^l; \; \mathcal{C}^l] \big), \quad
\mathbf{q}_i^{l+1} \leftarrow \mathbf{q}_i^{l+1} + f_{gs}^l,
\end{equation}
where $[\cdot; \cdot]$ denotes concatenation and $\mathbf{q}_i^{l+1}$ is the query feature for the $(l+1)$-th layer.

By informing deeper layers of what shallower layers have established, GSPE transforms decoding into residual refinement: early layers converge on coarse scene layout, providing stable anchors that guide fine-grained convergence of deeper layers.

\paragraph{Optimization Objectives.} 

To train our network end-to-end, we formulate a comprehensive multi-task loss function that jointly supervises semantic accuracy, geometric topology, and primitive spatial distribution. The total optimization objective $\mathcal{L}_{total}$ is defined as a weighted combination:
\begin{equation}
\label{eq:total_loss}
\mathcal{L}_{total} = \lambda_{nll} \mathcal{L}_{nll} + \lambda_{lov} \mathcal{L}_{lov} + \lambda_{bce} \mathcal{L}_{bce} + \lambda_{cd} \mathcal{L}_{cd},
\end{equation}
where $\lambda_{nll}$, $\lambda_{lov}$, $\lambda_{bce}$, and $\lambda_{cd}$ denote the balancing hyperparameters. 

For per-voxel semantic supervision, we employ the negative log likelihood (NLL) loss, which naturally couples with the probabilistic superposition formulation that outputs categorized probability distributions at each voxel. 

Following the convention of TPVFormer~\cite{huang2023tpvformer}, we further adopt the Lov\'asz loss $\mathcal{L}_{lov}$~\cite{berman2018lovasz} to refine geometric shape supervision at the voxel level.

For geometric shape supervision, we formulate a binary cross entropy loss $\mathcal{L}_{bce}$ over the occupancy state of each voxel. This term explicitly forces the synthesized Gaussians to conform to the exact geometric boundaries. The Chamfer Distance loss $\mathcal{L}_{cd}$ defined in Eq.~(\ref{eq:cdloss}) regularizes the spatial distribution of Gaussian centers, stabilizing primitive positions from a cold start.

\section{TJScenes Dataset}

\begin{figure*}[t]
    \centering
    \includegraphics[width=\textwidth]{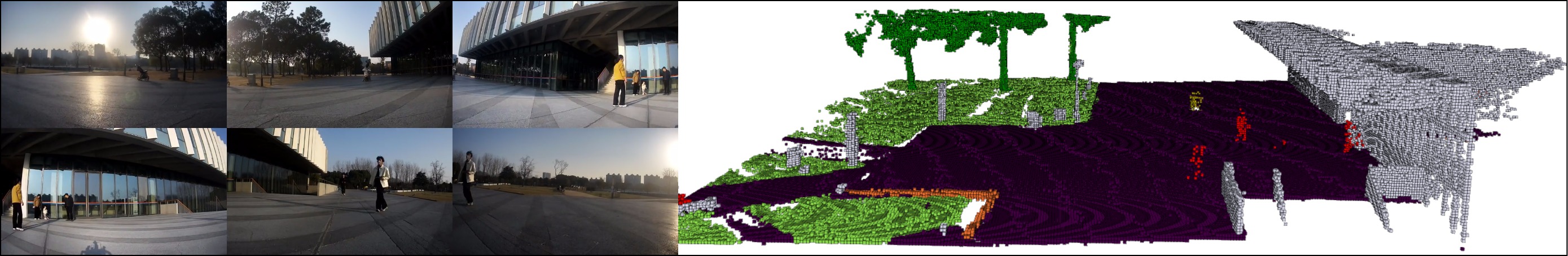}
    \caption{\textbf{Example panoramic images and associated ground truth voxels in TJScenes.} This representative sample captures a plaza-like off-road environment, highlighting a critical scenario that is largely absent from current mainstream autonomous driving datasets.}
    \label{fig:T03_voxels_offroad}
\end{figure*}

\begin{figure*}[t]
    \centering
    \includegraphics[width=\textwidth]{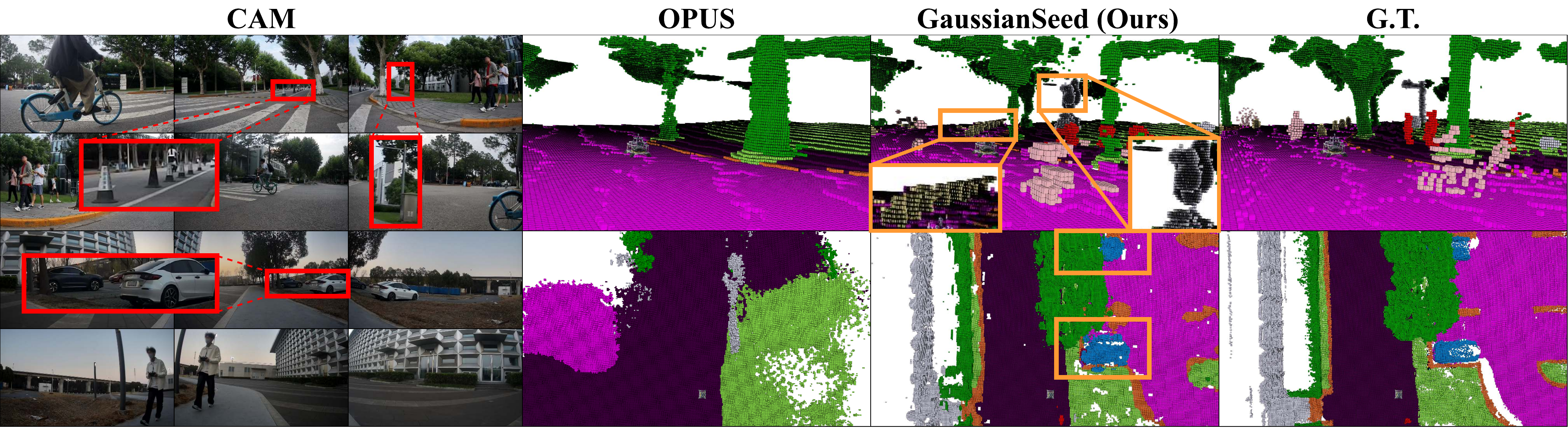}
    \caption{\textbf{Qualitative comparison on TJScenes.} Top row: at a crossroad, our method successfully captures small and distant objects such as traffic cones and streetlamps, and correctly detects dynamic objects like bicycles. Bottom row: in an off-road playground scenario, our method detects cars behind trees in the parking lot, whereas OPUS~\cite{wang2024opus} exhibits reduced robustness to off-road settings.}
    \label{fig:tjscenes_bench}
\end{figure*}

\begin{table*}[t]
\centering

\begin{tabular}{l l l c c c}
\toprule
Method & Venue & Backbone & Image Size & mIoU (\%) & Latency (ms) \\
\midrule
\multicolumn{6}{c}{\textbf{Dense Grid Representations}} \\
\midrule
BEVFormer~\cite{bevformer2022li} & ECCV'22 & ResNet-101 & $900 \times 1600$ & 39.30 & 213.2 \\
TPVFormer~\cite{huang2023tpvformer} & CVPR'23 & ResNet-101 & $928 \times 1600$ & 27.83 & 302.3 \\
FB-Occ~\cite{li2023fbocc}    & ICCV'23 & ResNet-50  & $256 \times 704$  & 37.50 & 128.4 \\
SurroundOcc~\cite{surroundocc2023Wei} & ICCV'23 & ResNet-101 & $800 \times 1333$ & 34.40 & 335.7 \\
PanoOcc~\cite{wang2024panoocc}   & CVPR'24 & ResNet-101 & $900 \times 1600$ & 42.10 & 335.1 \\
COTR~\cite{cotr2024Ma}      & CVPR'24 & ResNet-50  & $256 \times 704$  & \underline{44.50} & 1009.2 \\
ProtoOcc~\cite{protoocc2025kim}   & AAAI'25 & ResNet-50  & $256 \times 704$ & 39.56  & \underline{77.9} \\
GSD-Occ~\cite{gsdocc2025he}   & AAAI'25 & ResNet-50  & $512 \times 1408$ & 41.70 & 100.0 \\
ALOcc~\cite{gsdocc2025he}     & ICCV'25 & Intern-T   & $256 \times 704$  & \textbf{47.50} & \textbf{77.3} \\
\midrule
\multicolumn{6}{c}{\textbf{Fully Sparse Representations}} \\
\midrule
SparseOcc~\cite{liu2023sparseocc} & ECCV'24 & ResNet-50  & $256 \times 704$  & 30.60 & 80.2 \\
OPUS~\cite{wang2024opus}      & NeurIPS'24 & ResNet-50  & $256 \times 704$  & \underline{33.27} & \underline{45.1} \\
GaussianFormer~\cite{huang2024gsformer1} & ECCV'24 & ResNet-50  & $256 \times 704$  & 32.10 & 218.7 \\
\textbf{GaussianSeed (Ours)} & - & ResNet-50 & 256 $\times$ 704 & \textbf{34.12} & \textbf{41.2} \\
\bottomrule
\end{tabular}
\caption{\textbf{Comparison of 3D occupancy prediction methods on the Occ3D-nuScenes benchmark.} We categorize the methods based on their underlying spatial representation. While dense methods achieve higher mIoU, they suffer from significant latency. All latencies are measured on a single NVIDIA L40 GPU. Best results are \textbf{boldfaced} and second-best are \underline{underlined}.}
\label{tab:results_1_occ3d}
\end{table*}

While most existing benchmarks predominantly focus on standard vehicular road typologies, they often fall short for robotics applications, which demand a more granular understanding of non-driving layouts. Confronting these unique robotic operational demands, we collected TJScenes, a specialized dataset captured in dynamic campus environments tailored for micro-mobility and robotic navigation.

\begin{figure}[ht]
    \centering
    \includegraphics[width=\columnwidth]{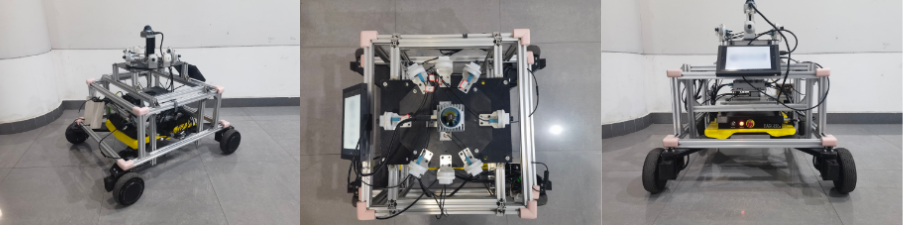}
    \caption{\textbf{The customized platform for data collection.}}
    \label{fig:agv}
\end{figure}

Specifically, TJScenes encompasses 7 distinct scenes, yielding a total of 27911 annotated samples. Each sample features synchronized images from a 6-camera array mounted on a customized mobile robotic platform (as shown in Fig.~\ref{fig:agv}) to ensure full $360^\circ$ surround-view coverage, accompanied by high-quality 3D semantic occupancy annotations, derived from Livox Mid-360 LiDAR scans, at a fine-grained $0.1\text{m}$ resolution spanning a $[\pm 20\text{m}, \pm 20\text{m}, -2\text{m} \sim 4.4\text{m}]$ perceptual range.

\begin{figure}[ht]
    \centering
    \includegraphics[width=\columnwidth]{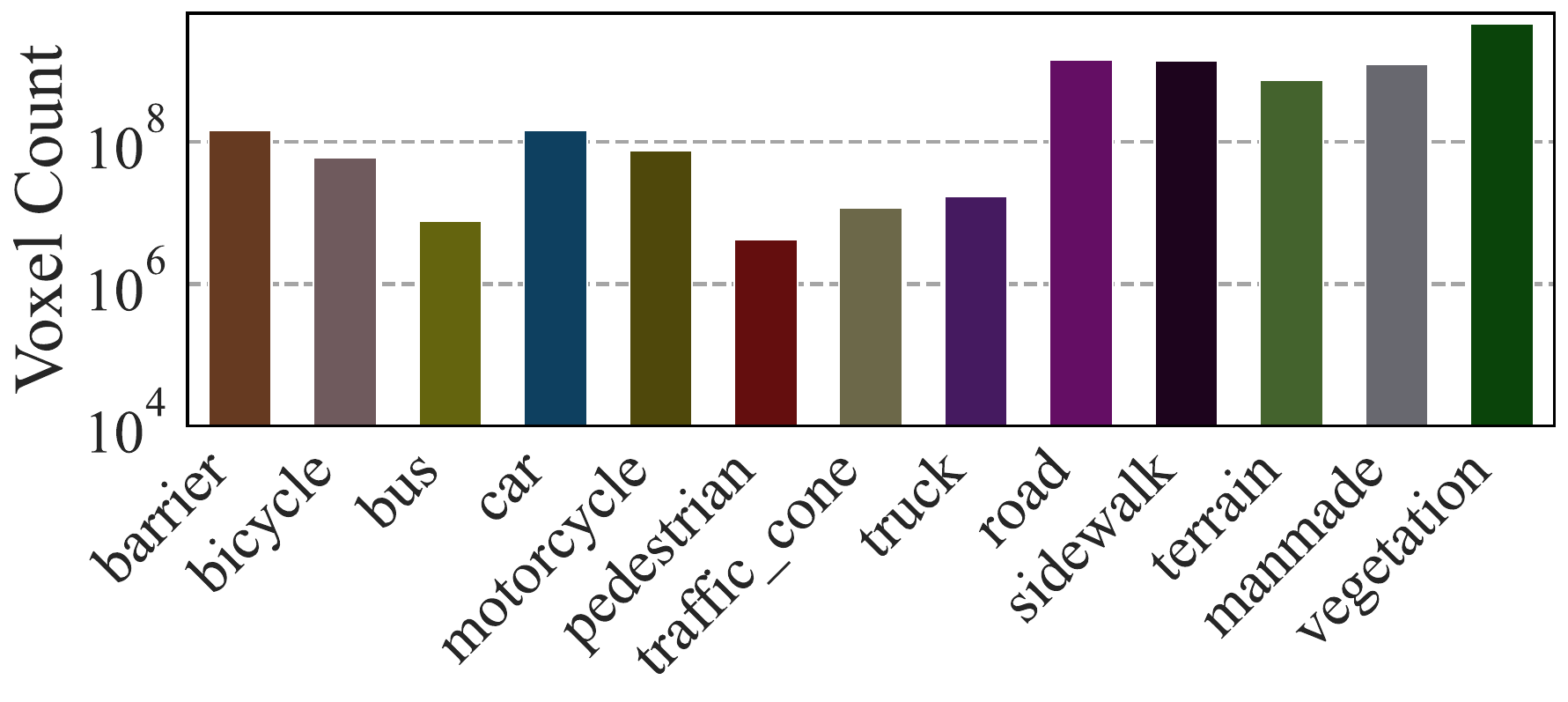}
    \caption{\textbf{Class distribution of the TJScenes dataset.} The dataset encompasses a diverse spectrum of dynamic objects and static background structures.}
    \label{fig:cls_dis}
\end{figure}

TJScenes uniquely incorporates diverse \textit{sidewalk} and \textit{off-road} scenarios that are rarely emphasized in highway-centric datasets. The dataset features comprehensive annotations across 13 standard semantic categories, as illustrated in Fig.~\ref{fig:cls_dis}. Crucially, navigating these narrow, unstructured campus paths places a significantly higher premium on precise boundary delineation, making TJScenes an ideal testbed to validate the capability of GaussianSeed in resolving $0.1\text{m}$ high-fidelity geometric structures.

\section{Experiments}


\paragraph{Datasets and Metrics.}

We evaluate on both Occ3D-nuScenes and TJScenes as described earlier, and adopt mean Intersection-over-Union (mIoU) as the primary metric, following the standard convention.

\paragraph{Implementation Details.}

We adopt ResNet-50 as our image backbone to maintain training efficiency and moderate inference latency. For input specifications, camera images are resized and cropped to $704 \times 256$ for both datasets. We use the current frame together with 7 preceding frames (8 frames in total) as temporal input.
The entire framework is optimized end-to-end using the AdamW optimizer with a weight decay of $0.01$. We employ a cosine annealing learning rate scheduler with an initial learning rate of $2 \times 10^{-4}$, preceded by a linear warmup phase of $500$ iterations. For all experiments, the total batch size is set to $32$ and distributed across 8 NVIDIA L40 GPUs.

\paragraph{Main Results.}

\begin{figure*}[ht]
    \centering
    \includegraphics[width=\textwidth]{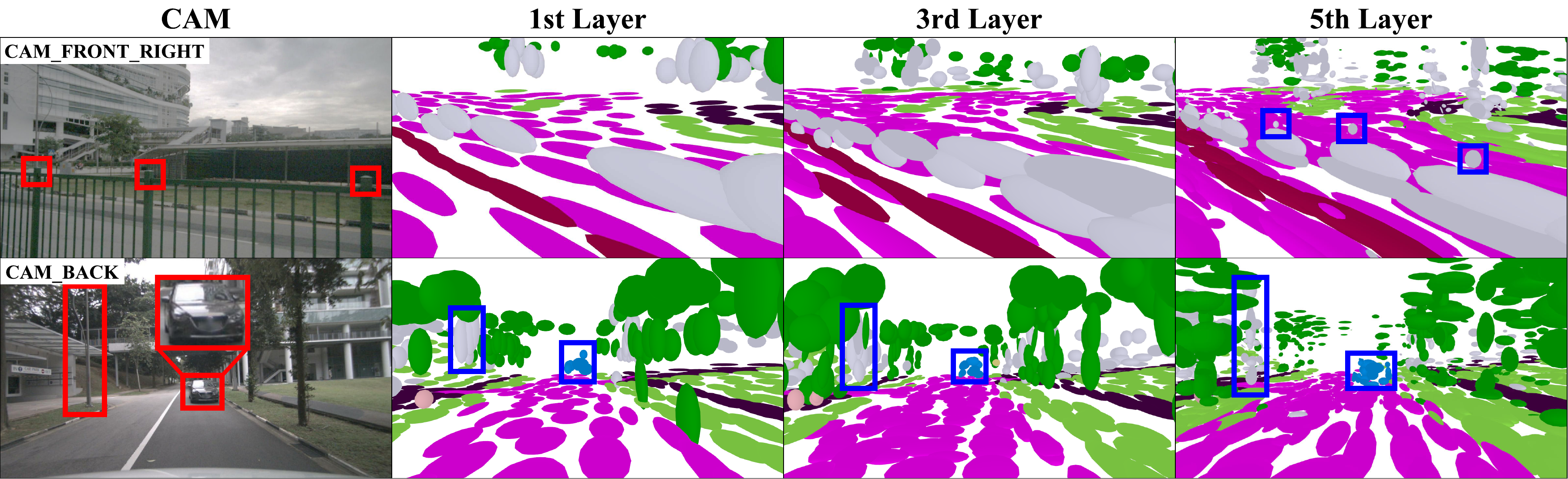}
    \caption{\textbf{Visualizing the hierarchical evolution of HGSD.} From left to right, we observe the progressive refinement of Gaussian primitives. Early stages construct the rough geometric layout of a vehicle, while subsequent stages devolve into finer, highly-adapted ellipsoids to capture intricate surface details.}
    \label{fig:hgsd_qualititive_analysis}
\end{figure*}

\begin{table*}[t]
\centering

\begin{tabular}{l l l c c c c c}
\toprule
Method & Backbone & Image Size & mIoU (\%)&  mIoU$_D$ (\%) & Memory (MB) & Latency (ms) \\
\midrule
OPUS ($N_{\text{Queries}}=800$)      & ResNet-50  & $256 \times 704$  & 7.87 & 5.37 & 1168.01 & 97.8 \\
\textbf{GaussianSeed (Ours)} & ResNet-50 & 256 $\times$ 704 & \textbf{12.37} & \textbf{7.60} & \textbf{931.93} & \textbf{66.7} \\
\bottomrule
\end{tabular}
\caption{\textbf{Quantitative results on TJScenes at 0.1m resolution.} The evaluation metrics $\text{mIoU}_D$ represent the average performance across dynamic categories, including pedestrian, car, bicycle, motorcycle, bus, truck.}
\label{tab:results_2_tjscenes}
\end{table*}

We evaluate our framework against both dense and sparse baselines. As shown in Tab.~\ref{tab:results_1_occ3d}, dense architectures push the performance upper bound, but this comes at the cost of severe computational overhead, with latencies typically exceeding 100ms. Sparse methods provide a more efficient alternative. Among them, our GaussianSeed achieves the best accuracy (34.12\% mIoU) and the lowest latency (41.2ms) simultaneously, making it the top-performing sparse method across both dimensions.

Although our primary design objective is to enable scaling to high-resolution spatial predictions, the fully sparse nature of our architecture inherently minimizes latency. Specifically, GaussianSeed achieves an average inference time of 41.2ms, making it the fastest framework among all evaluated methods. 

Notably, benefiting from the hierarchical design, our method simultaneously improves accuracy over GaussianFormer while reducing latency by more than $5\times$, demonstrating that Gaussian-based occupancy representations can be deployed for real-time inference.

The true scaling capability of GaussianSeed is best demonstrated on the much more demanding TJScenes dataset, as shown in Tab.~\ref{tab:results_2_tjscenes}. At the resolution of $0.1 \text{m}$, memory limits become a critical barrier for previous methods. Scaling to such a dense space causes all the aforementioned dense baselines to experience memory exhaustion, unless forced into heavy parameter tuning that ultimately degrades their accuracy. As a result, OPUS remains the only baseline capable of native training on this dataset. While our framework matches the performance of OPUS in reconstructing static backgrounds, it shows a significant advantage when predicting dynamic objects. This improvement is directly tied to the anisotropic properties of 3D Gaussians. Unlike the discrete point queries used in OPUS, Gaussian primitives can independently optimize their scale and rotation. This allows them to stretch and tightly wrap around the complex boundaries of vehicles and pedestrians, capturing local geometric details much more effectively. GaussianFormer series are not included in this benchmark, since GaussianFormer suffers from intractable memory explosions on our $0.1\text{m}$ dense grid due to empty-space modeling, while GaussianFormer-2's reliance on a highly coupled, dataset-specific distribution initialization pipeline prevents its fair generalization to our novel resolution.

\paragraph{Visualizations.}

While deep networks are often regarded as black boxes, the hierarchical devolution process of HGSD offers interpretable physical behaviors. To demystify this, we visualize the step-by-step deformation of Gaussian primitives across different devolution stages in Fig.~\ref{fig:hgsd_qualititive_analysis}. For structured objects, the initial devolution stages establish a coarse skeleton. As the primitives propagate to subsequent deeper layers, they undergo finer local splitting and elongation, manifesting as more compact, highly-oriented Gaussian ellipsoids that precisely wrap around thin surfaces. This empirical observation confirms that HGSD successfully guides the network to learn a coarse-to-fine geometric representation.

This structural advantage is visually illustrated in Fig.~\ref{fig:tjscenes_bench}. Despite occasional ghosting artifacts on several dynamic objects, the Gaussian primitives still show a clear advantage in capturing the local geometries of small scale objects. While OPUS's discrete point queries occasionally lead to fragmented or overly coarse shapes, our primitives dynamically adapt their scales and orientations to wrap more tightly around complex boundaries.

\paragraph{Ablation Studies.}

We conduct ablation studies to validate each core component of GaussianSeed. All ablations use ResNet-50 with $256 \times 704$ input resolution under identical training protocols.

\noindent\textbf{Architecture design.}
We evaluate the contributions of Hierarchical Gaussian Seed Devolution (HGSD) and the Gaussian Seed Parameter Encoder (GSPE) by enabling each module independently. As shown in Tab.~\ref{tab:ablation_arch}, both components bring consistent gains over the plain baseline on both Occ3D-nuScenes and TJScenes ($0.1\text{m}$). HGSD contributes larger gains by structuring prediction into a coarse-to-fine curriculum, while GSPE adds complementary benefit through explicit cross-layer feature propagation. Their combination achieves the best accuracy across both benchmarks.

\begin{table}[t]
\centering
\begin{tabular}{c c c c c}
\toprule
HGSD & GSPE & \makecell{Occ3D\\ mIoU (\%)} & \makecell{TJScenes\\ mIoU (\%)} \\
\midrule
\ding{55} & \ding{55} & 25.81 & 6.37 \\
\ding{51} & \ding{55} & 32.47 & 11.52 \\
\ding{55} & \ding{51} & 26.79 & 7.24 \\
\ding{51} & \ding{51} & \textbf{34.12} & \textbf{12.36} \\
\bottomrule
\end{tabular}
\caption{Architecture ablation. \ding{51}: enabled; \ding{55}: disabled.}
\label{tab:ablation_arch}
\end{table}

\noindent\textbf{Gaussian count.}
We further investigate the effect of Gaussian primitive allocation under a fixed query budget. We vary the total Gaussian count $G_{\text{total}}$ while keeping the number of queries at $Q = 800$, and conversely vary $Q$ while fixing $G_{\text{total}} = 28800$. As shown in Tab.~\ref{tab:ablation_gscount}, performance on Occ3D-nuScenes initially improves with more Gaussians per query, peaking at $28800$, after which further increases yield diminishing returns. Conversely, keeping $G_{\text{total}}$ fixed while varying $Q$ reveals that $Q = 800$ achieves the best accuracy. Too few queries coarsen spatial coverage, while too many dilute the per-query Gaussian budget and weaken local expressiveness. Notably, under the 600-query setting matching OPUS-T, our method still slightly outperforms it (33.71\% vs.\ 33.27\%), further confirming the efficiency of our hierarchical design.

\begin{table}[t]
\centering
\begin{tabular}{c c c}
\toprule
Queries $Q$ & Total Gaussians $G_{\text{total}}$ & mIoU (\%) \\
\midrule
800  & 7200   & 27.53 \\
800  & 14400  & 31.47 \\
800  & \textbf{28800} & \textbf{34.12} \\
800  & 43200  & 33.68 \\
600  & 28800  & 33.71 \\
1000 & 28800  & 33.24 \\
\bottomrule
\end{tabular}
\caption{Gaussian count ablation on Occ3D-nuScenes. Default setting in bold.}
\label{tab:ablation_gscount}
\end{table}

\section{Conclusion}

In this paper, we presented GaussianSeed, an efficient hierarchical framework for 3D semantic occupancy prediction. By introducing RBGI, our method elegantly achieves robust Gaussian initialization without relying on any explicit depth supervision. Furthermore, driven by a progressive coarse-to-fine hierarchical design, GaussianSeed successfully scales to the highly demanding $0.1\text{m}$ spatial resolution. Extensive evaluations demonstrate that our approach sets a new efficiency-quality frontier, delivering the lowest inference latency while maintaining highly competitive accuracy on Occ3D-nuScenes. Most notably, on the fine-grained TJScenes benchmark, GaussianSeed significantly outperforms existing sparse paradigms, underscoring its exceptional capability and practical value for intricate geometric perception in micro-mobility and robotic navigation.

\bibliography{ref.bib}


\end{document}